\documentclass[11pt,a4paper]{article}
\usepackage[hyperref]{acl2019}
\usepackage{times}
\usepackage{latexsym}
\usepackage{CJKutf8} 
\usepackage{url} 
\usepackage{graphicx} 
\usepackage{subfig} 
\usepackage{algorithm}
\usepackage{algorithmic} 
\usepackage{amsmath,amssymb,amsfonts} 
\usepackage{multirow} 

\aclfinalcopy 

\title{A Seq-to-Seq Transformer Premised Temporal Convolutional Network for Chinese Word Segmentation}

\author{Wei Jiang \textnormal{and} Yan Tang\thanks{* Corresponding author.} \\
  School of Computer and Information Science,Southwest University \\
  No 2,Tiansheng Road,Chongqing,China \\
  \texttt{jw2312@email.swu.edu.cn} \\
  \texttt{ytang@swu.edu.cn} \\}

\date{}

\begin{document}
\maketitle
\begin{abstract}
  The prevalent approaches of Chinese word segmentation task almost rely on 
  the Bi-LSTM neural network. However, the methods based the Bi-LSTM have some inherent drawbacks: hard to parallel computing, little efficient in applying the Dropout method to inhibit the Overfitting and little efficient in capturing the character information at the more distant site of a long sentence for the word segmentation task. In this work, we propose a sequence-to-sequence transformer model for Chinese word segmentation, which is premised a type of convolutional neural network named temporal convolutional network. The model uses the temporal convolutional network to construct an encoder, and uses one layer of fully-connected neural network to build a decoder, and applies the Dropout method to inhibit the Overfitting, and captures the character information at the distant site of a sentence by adding the layers of the encoder, and binds Conditional Random Fields model to train parameters, and uses the Viterbi algorithm to infer the final result of the Chinese word segmentation. The experiments on traditional Chinese corpora and simplified Chinese corpora show that the performance of Chinese word segmentation of the model is equivalent to the performance of the methods based the Bi-LSTM, and the model has a tremendous growth in parallel computing than the models based the Bi-LSTM. 
\end{abstract}

\section{Introduction}

  Chinese word segmentation (CWS) is a preliminary task for Natural Language Processing (NLP) of Chinese. The CWS has regarded as a labeling problem since Xue~\shortcite{Xue:2003}. Previous research of the CWS focussed on statistical methods based on supervised machine learning algorithms, such as Maximum Entropy~\cite{Berger:1996} and Conditional Random Fields~\cite{lafferty:2001}. However, those methods heavily depend on the selecting of handcrafted features.
  
  Recently, the neural network models have widely adhered to solving the NLP tasks for their ability to minimize the effort in feature engineering. The research attention in the CWS has shifted to deep-learning~\cite{Zheng:2013,Pei:2014,Chen:2015,Cai:2016,Chen:2017,Yang:2017,Ma:2018}. Deep-learning practitioners commonly regard recurrent architectures as a default starting point for sequence modeling tasks. Chen~\shortcite{Chen:2015} built a recurrent neural network (RNN) for the CWS and applied Long short-term memory (LSTM)~\cite{Hochreiter:1997} as the cell operator of the RNN. Cai~\shortcite{Cai:2016} employed gate combination neural network and LSTM to establish a word-based model of the CWS. Ma~\shortcite{Ma:2018} Proposed a relatively simple and efficient model by stacking forward and backward LSTM cell chains, which is one of the best models based the Bi-LSTM for the CWS. 

  However, the methods based on the Bi-LSTM have some inherent drawbacks. First, they are hard to parallel computing for the correlation that the computing of each hidden state of the Bi-LSTM needs the value of the previous hidden state. Second, they are little efficient in applying the Dropout~\cite{Hinton:2014} method to inhibit the Overfitting for the correlation. Third, for the word segmentation task, they are little efficient in capturing the character information at the more distant site of a long sentence for the vanishing gradients. Contrasted the Bi-LSTM, Convolutional Neural Network (CNN) has more advantages in parallel computing and extraction of features.  
  
  It is difficult that using the traditional convolutional neural network builds a sequence model. For language modeling task, the extraction of the information of context is a key obstacle for using the traditional convolutional neural network. Kim~\shortcite{Kim:2014} used CNN to extract features from character-based embeddings and concatenate the features as the input of the LSTM neural network that predicted the next word in a sentence. Gehring~\cite{Gehring:2017a,Gehring:2017b} proposed an architecture composed of an encoder and a decoder for machine translation that the encoder was a primitive temporal convolutional network. Bai~\shortcite{Bai:2018} proposed a generic temporal convolutional network architecture for numerous tasks, which was a systematical description of the temporal convolutional network firstly. However, The temporal convolutional network that Bai proposed is weak in the tasks of language modeling liking the CWS. 
  
  In this paper, we propose a sequence-to-sequence transformer model for the task of Chinese word segmentation. The base of the model is temporal convolutional network. The temporal convolutional network, which derives from Bai's model and is improved by us, mends the shortcomings of Bai's model and adheres to the task of Chinese word segmentation. The model uses the temporal convolutional network to construct an encoder and uses one layer of fully-connected neural network to build a decoder. Meanwhile, the model applies the Dropout method to inhibit the Overfitting and captures the character information at the distant site of a sentence by adding the layers of the encoder. The most key points are that the model binds the Conditional Random Fields model to train parameters, and the model uses the Viterbi algorithm to infer the final result of the Chinese word segmentation. For further research, we release the data and the source code publicly \footnote{\texttt{https://github.com/Johnwei386/Tcn\_CWS}}.

  In summary, the contributions of this paper are concluded as follows: 

\begin{itemize}
\item We firstly take a tentative that we introduce the temporal convolutional network to solve the task of Chinese word segmentation. 
\item We improve the generic temporal convolutional network (TCN) in three aspects, architecture, training, inference. It makes the model sticking to the task of Chinese word segmentation. The experiments in performance of the CWS will show that our model achieves the best result and a tremendous growth contrasted the generic temporal convolutional network. 
\end{itemize}  

\section{The Transformer Model}
\label{sec:model}

\begin{figure*}[!t]
\centering
\includegraphics[width=6.0in]{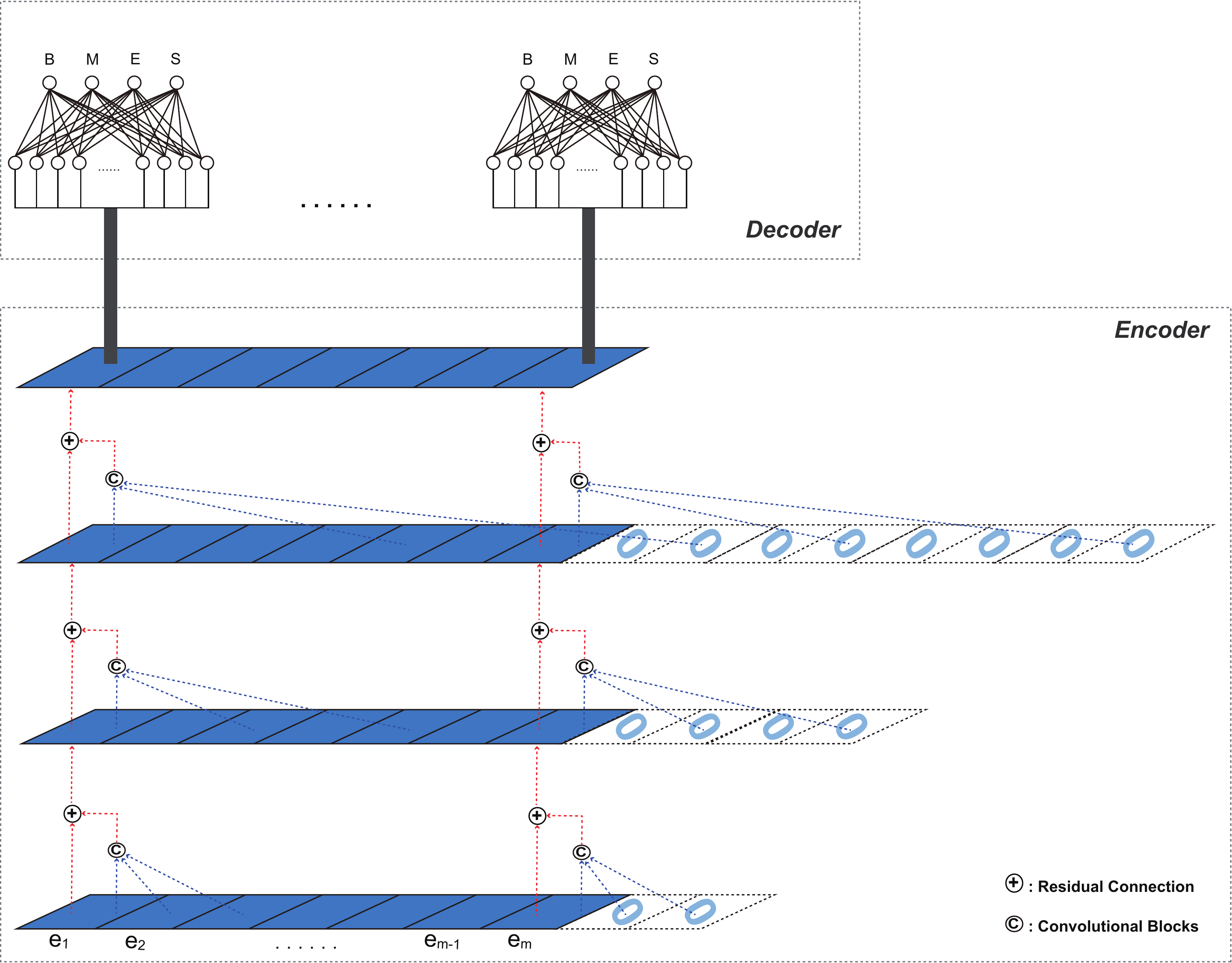}
\label{edArch}
\caption{The architecture of Seq-to-Seq transformer. The bottom layer of the encoder is the input layer. A blue dotted line indicates the capturing of a receptive field. A blue block encompassed by the black solid line is an element of layers. A zero block that contains a zero is a zero padding block. The dilation factors for hidden layers are set as $d = 2^0,2^1,2^2$.
}
\label{tcnArch}
\end{figure*}
 
The transformer model that we propose, as shown in Figure~\ref{tcnArch}, consists of an encoder and a decoder. The encoder is a temporal convolutional network. The temporal convolutional network is composed of an input layer, many hidden layers, and an output layer. The hidden layers are dilated convolutional layers that are the results after performing the operations of \textbf{Convolutional Blocks} and \textbf{Residual Connection} to the inputs from previous layers. The dilated convolutional layers have a mechanism named  \textbf{dilated convolution} for capturing the receptive field. The convolutional neural network has two important concepts, receptive field and filter. A receptive field is a scope of artificial neurons for a convolutional operator in CNN. For example, in our model, the receptive field of a convolutional operator of the first hidden layer is a range of characters or words in a sentence. If the elements of the receptive field are successive in the previous layer, the convolutional operator is a regular convolution, otherwise, it is a dilated convolution. When the elements of the receptive field are captured from the previous layer, a picked element has an interval with another picked element that they are both in the previous layer. It is the core idea of the dilated convolution. In general, the interval is a hyper-parameter named \textbf{dilation factor} that is set as an exponent of two generally. The input of each dilated convolutional layer is a concatenation by numerous receptive fields of dilated convolution operators binding the elements of this layer.

Chinese word segmentation task is usually regarded as a sequence labeling task by characters. Specifically, each character in a Chinese sentence is classified as a label of $L = \{B,M,E,S\}$, indicating the begin site of a word, the middle site of a word, the end site of a word, or a word with a character. A sentence with m characters can be described as $X = \{ x_1,x_2,\ldots,x_m \}$, and the aim of the CWS is to figure out the a true label sequence that can be described as $Y = \{ y_1,y_2,\ldots, y_m \}, y_t \in L, t = 1,2,\ldots,m$, $t$ indicating an index of characters in the sentence. The original $X$ cannot be applied directly to our model. It is necessary to transform $X$ to pre-trained embedding vectors. The pre-trained embedding vectors can be described as $E = \{e_1, e_2, \ldots, e_m\}, e_t \in R^n$, the $e_t$ binding the $x_t$, which is the input layer of the temporal convolutional network. $n$ is a hyper-parameter indicating the dimensions of a pre-trained embedding vector. 

\subsection{The Encoder}
\label{sect:encoder}
The encoder is a convertor that converts the input layer to the output layer. The input layer is a concatenation of pre-trained embedding vectors that it indicates a sentence. For the task of Chinese word segmentation, it is necessary to keep the length of the input layer equivalent to the length of the output layer. It is an alignment mechanism between the input layer and the output layer. It also means that every character of the sentence inputted to the encoder binds the element of the output layer with the same index.

The hidden layers gain receptive fields by the dilated convolution, each receptive field binding an element of the hidden layer where the dilated convolution is performed. As the reason of the alignment mechanism, each element of the hidden layers at the same index with the character of the sentence inputted to the encoder has a receptive field. It is equivalent to that a convolutional filter slides on the hidden layers from left to right with the stride equalling 1 to cruising the elements one by one. A Convolutional filter is an extractor of features, which has a set of independent parameters. When an element of a hidden layer gained the receptive field, an operation of the Convolutional Blocks on the receptive field is performed to generate a result, and an operation of Residual Connection is performed with the result and the element of the previous layer at the same index with the element of the hidden layer to generate the output of the element of the hidden layer. The output of the last hidden layer is the output layer.

A improvement of our model in architecture is that a receptive field of a hidden layer is captured only from an element to the elements at the right direction in the previous layer that the element has the same index with the spot in the hidden layer where a convolutional filter is staying. It is named \textbf{future scheme}. Instead, the scheme that only focusses the capturing of the elements at the left direction is named \textbf{past scheme}.

\subsection{Dilated convolution}
\label{sect:dilatedconv}
The dilated convolution gains the elements that a receptive field needs. For a sentence $X$  inputted to the encoder, the elements can be described as a set of $x_{t+d \cdot i}$, $t$ indicating the site where the convolutional filter is staying, $d$ indicating the dilation factor, $i$ indicating an index of the receptive field. The dilation factor of each hidden layer is an exponent of 2, but the power of the exponent is incremental that it is different for each hidden layer. 

The dilated convolution causes that the receptive fields of the last elements of hidden layers cannot capture sufficient elements of previous layers, So, it is necessary adding zero padding blocks to the tail of the layers. A zero padding block also is the element of the layers but without the receptive field, and the value of the zero padding block is zero. As the layers of the encoder adding, the transformer model can capture the character information at the distant site of a sentence by the dilated convolution.

\subsection{Convolutional Blocks}
\label{sect:convblocks}

\begin{figure}[!t]
\centering
\includegraphics[width=2.5in]{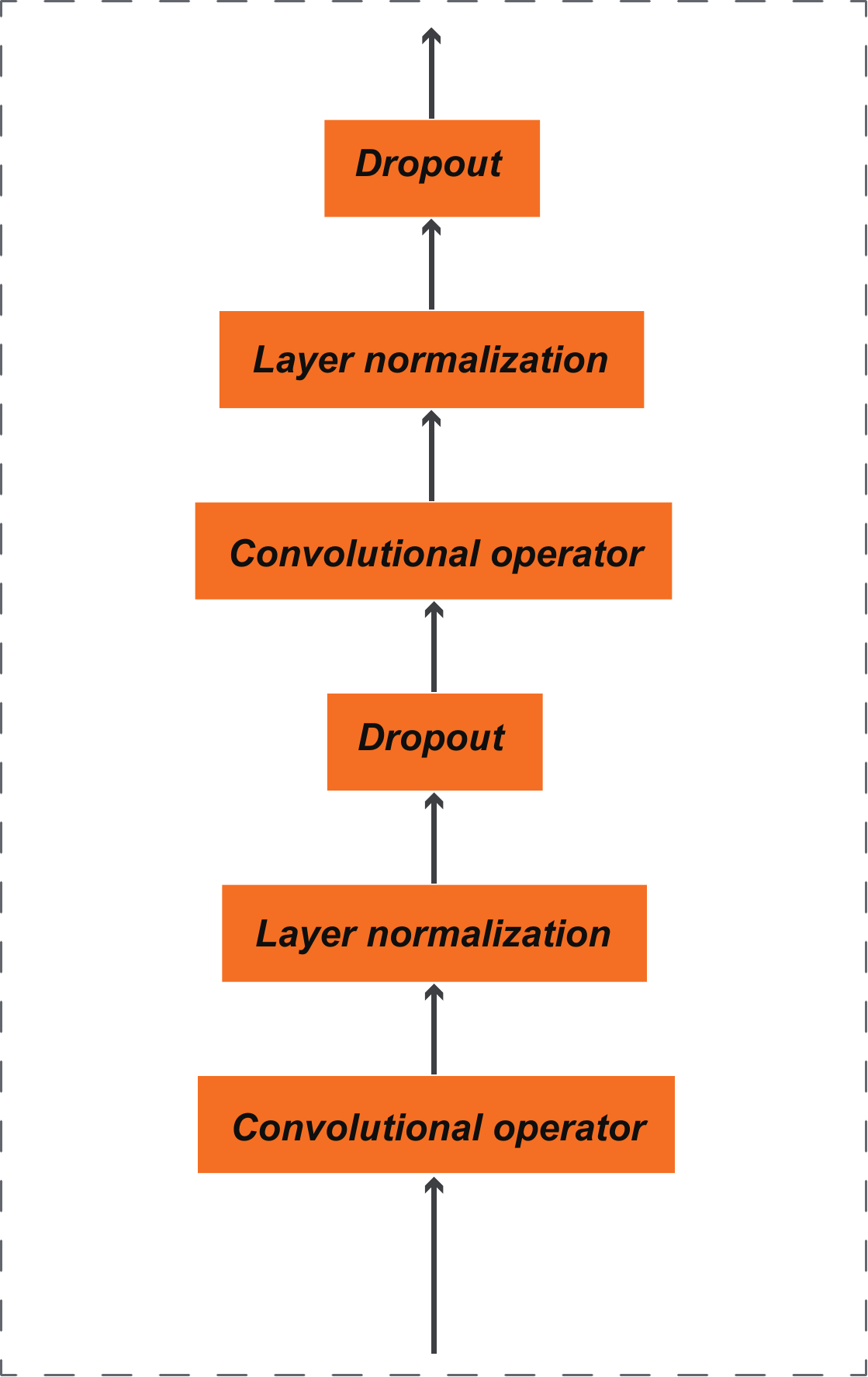}
\caption{The Convolutional Blocks.}
\label{convOperator}
\end{figure}

A convolutional block is composed of three transformations for the elements of a receptive field. The transformations are Convolutional Operator, Layer Normalization, Dropout. Numerous convolutional blocks are stacked as an integral operation named  \textbf{Convolutional Blocks}. In our model, as shown in Figure~\ref{convOperator}, we only stack two convolutional blocks.

The Convolutional Operator is the traditional convolutional operator of one-dimension in CNN. The Layer Normalization is a simple normalization method to improve the training speed for various neural network models proposed by Jimmy~\cite{Jimmy:2016}, which is designed to overcome the drawbacks of batch normalization~\cite{Sergey:2015}. The Dropout is a prevalent method to avoid the Overfitting for numerous neural networks.

\subsection{Residual Connection}
\label{sect:residual}
The Residual Connection is a connection way between different layers of a neural network, which was first applied in computer vision study. Traditionally, The input of a layer of the neural network is the output of the previous layer directly. However, by the Residual Connection, The input of a layer of the neural network is the sum of the primitive output after all operations of the previous layer and the input of the previous layer. In our model, an output of an element of a hidden layer is described as:

\begin{equation}
\label{eq1}
output = ReLU(ve + cblocks)
\end{equation}

In Eq~\ref{eq1}, we use ReLU as the activation function, and $ve$ is the value of the element of the previous layer that has the same index with the character of the sentence where a convolutional filter is processing, and $cblocks$ is the output of the Convolutional Blocks for the element, and $output$ is the final output for the element. The concatenation of all $output$ is the output of the hidden layer.

\subsection{The Decoder}
\label{ssec:decoder}
When the encoder processed a concatenation of pre-trained embedding vectors that indicates a sentence, the output layer of the encoder is a concatenation of vectors that the dimensions of the vectors are equivalent to pre-trained embedding vectors. The decoder is constructed by one layer of fully-connected neural network, and it is also a convertor that converts the vector of the output layer to the vector contained the score of each label that is one of  the classification labels of the CWS. The dimensions of the vector contained the score are set as 4, indicating the four classifications for the CWS.

\section{Training}
\label{sec:training}

\begin{figure}[!t]
\centering
\includegraphics[width=2.5in]{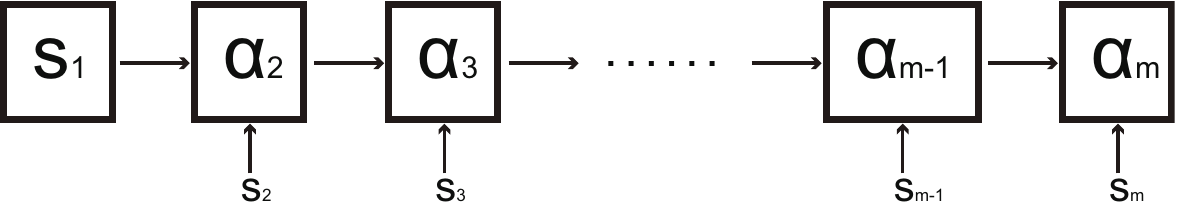}
\caption{The computing precedure of $\alpha$.}
\label{fig:alpha}
\end{figure}

We bind the Conditional Random Fields (CRF) to train our model. The CRF is a simple and effective model for the task of sequence labelling. The idea of the CRF is that seeing adjacent labels in a sequence is dependent instead of independent. In common, the CRF has a transform matrix indicating the weight to transfer between the adjacent labels and holds two paths at forward-path and backward-path for a sequence. In our model, for simple in practice, we only consider the interaction between two successive labels at forward-path. For a training sample to our model $(x_t, y_t)^m$, where $y_t$ is the true label for $x_t$ and $m$ is the length of the sentence inputted to the encoder, the $\hat{Y}$, the probability for all true labels in the sentence, is designed as: 

\begin{small}
\begin{equation}
\label{eq2}
\hat{Y} = \frac{exp\left(\sum_{t=1}^m \left( score_t(y_t) + \varphi (y_{t-1},y_t)\right)\right)}{\sum\limits_{l \in L} exp(\alpha_m (l))}
\end{equation}
\end{small}

In Eq~\ref{eq2}, The $score$ is the result of the the decoder, and $score_t(y_t)$ means the score with true label at the index step $t$. The $\varphi$ is the function indicating the probability from the label of the previous step to transfer next one and $\varphi (y_0, y_1)$ is zero, which means for the first step without the dependent relationship, and $\varphi (r, c)$ is the mapping to the transform matrix that $r$ indicates the row of the matrix and $c$ indicates the column of the matrix.

It is hard to sum over an exponential number of state sequences for the CRF because it requires to compute the score based on the whole sentence. The score is the sum over all possible labeling schemes. The \textit{Forword-Backward} algorithm is a key solution to compute the score~\cite{MiCollins}, where $\alpha$ indicates the score at the forward direction. In our model, we only compute the score at forward-path. It is an iterative procedure to compute the $\alpha$ value. The $\alpha$ value at step $t$ for the label $l$ is described as:

\begin{small}
\begin{equation}
\label{eq3}
\alpha_t(l) = score_t(l) + log \sum\limits_{l^{'} \in L} exp(\alpha_{t-1}(l^{'}) + \varphi (l^{'}, l))
\end{equation}
\end{small}

The initial value for $\alpha_1(l)$ at the first step is equal to $score_1(l)$. The $\alpha_m$ means the score at the last index step $m$, which is the final score and describes the whole sentence. The procedure to compute $\alpha$ is shown at Figure~\ref{fig:alpha} and $s_t$ is the score at index step $t$ for a character, as well $s_1$ equaling to $\alpha_1$. 

We employ the Adam optimizer with $\beta_1 = 0.9$, $\beta_2 = 0.999$ and $\epsilon = 10^{-8}$ to optimize the model parameters~\cite{Kingma:2014}, which can get training away from plateau. Then, we use cross-entropy loss to train the model, and the loss function for a sentence is designed as:

\begin{equation}
\label{eq4}
loss = -log\left( \hat{Y} \right)
\end{equation}

\section{Inference}

\begin{algorithm}[!t]
\caption{The Viterbi algorithm}
\label{alg:viterbi}
\begin{algorithmic}
\REQUIRE ~~\\ 
The set of score, $score$; \\
The transform matrix, $M$; \\
\ENSURE ~~\\ 
\STATE $T_1 = score_1$
\FOR{$i=2$ to $m$}
\STATE $V_i = T_{i-1} \cdot I + M$
\FOR{$c=1$ to $\tau$}
\STATE $b_i(c) = \mathop{argmax}\limits_{k} V_i[k,c]$
\STATE $T_{i}(c) = score_{i}(c) + \left[\mathop{max}\limits_{k} V_i[k,c]\right]^T$
\ENDFOR
\ENDFOR
\STATE $w_m = argmax(T_m)$
\FOR{$j=m$ to $2$}
\STATE $w_{j-1} = b_j(w_j)$
\ENDFOR
\RETURN $W = \{w_1, w_2, \ldots, w_m\}$
\end{algorithmic}
\end{algorithm}

In common, it is a simple way to infer the labels by directly hitting the one that has the maximum score value label in the score for each character in a sentence. However, this way is not to take the interaction between two successive labels into account. So, we use the \textit{Viterbi algorithm} to infer the labels that are a true label sequence $Y$. It is a dynamic programming algorithm. 

As shown in Algorithm~\ref{alg:viterbi}, the $M$ is the transform matrix indicating the interaction between two adjacent tags, $M \in R^{\tau \times \tau}$, and $\tau$ indicates the size of label sets $L$, and $m$ is the size of a sentence, and $score_1$ is the score for each labels at first character in a sentence, $score_1 \in R^{\tau \times 1}$, and $T_i$ is a intermediate variable, and $V_i$ is a intermediary matrix with the row index $k$ and the column index $c$, $V_i \in R^{\tau \times \tau}$, and $I$ is a vector, $I = [1,1,\ldots,1], I \in R^{1 \times \tau}$, and $w_i$ indicates the inferred label result for the character, and $b_i$ indicates the row index having maximum value at column $c$ in matrix $V_i$, $b_i \in R^{1 \times \tau}$. The return of the \textit{Viterbi algorithm} is a serial $w_i$ as well as $W$. 

\section{Experiments}

\subsection{Datasets}
\label{ssec:datasets}
We use the NLPCC2016, a shared task for Weibo segmentation, as our main dataset~\cite{Qiu:2016}. It is split as Training, Development and Test set. In order to evaluate our proposed architecture, we additionally experiment on two prevalent CWS datasets, CITYU and PKU, which both derive from SIGHAN 2005 bake-off~\cite{Emerson:2005}, where the standard split is used. CITYU is a traditional Chinese corpus. NLPCC2016 and PKU are simplified Chinese corpora. For generating the pre-trained embedding vectors, We apply \textbf{wang2vec}\footnote{https:github.com/wlin12/wang2vec}~\cite{Ling:2015}, to pre-train the character-based embedding vectors. 

\subsection{Hyper-parameters}
\label{ssec:hyperparameters}

\begin{table}[t!]
\begin{center}
\begin{tabular}{l|c}
\hline 
\textbf{Paramter Description} & \textbf{Value} \\ 
\hline
Dimensions of embedding vector & $n = 100$ \\
Learn rate & $lr = 0.001$ \\
Filters number & $fs = 100$ \\
Hidden layers & $ly = 4$ \\
Kernel size & $s = 3$ \\
Dropout rate & $dp = 0.3$ \\
Dilation factors & $d = 2^i$ \\
Stride length & $sl = 1$ \\
Epochs & $ep = 100$ \\
Batch size & $bs = 32$ \\
\hline
\end{tabular}
\end{center}
\caption{The Hyper-parameters}
\label{hyper-parameters}
\end{table}

Hyper-parameters of neural network model significantly impacts its performance. To train our model and get a set of suitable hyper-parameters, we divide the training data into two sets, training set and development set. The development set contains 2000 sentence for each corpus experimented, and the rest of the corpus is the training set. The hyper-parameters of our model are shown in Table~\ref{hyper-parameters}. For the dropout rate, it is only available in the training procedure but has no impact in the test procedure. The kernel size and the dilation rate are set by commonly configuration policy. The stride length is the stride of a convolutional filter sliding to next spot, and it is only set as 1. Other hyper-parameters are trained by hand-manipulated.

\subsection{Experiment Environment and Evaluation Criteria}
The hardware environment is composed of an Intel Core i5 3.2GHz CPU with 8GB RAM and an Nvidia GeForce GTX 1050 Ti GPU. The software environment is composed of the Linux operating system, Tensorflow, Numpy. We apply the \textbf{F1} score to evaluate the performance of the CWS, which can be described as an \textbf{F} score. The F1 score is the harmonic average of the precision (\textbf{P}) and recall (\textbf{R}), where an F1 score reaches its best value at 1 (perfect precision and recall) and worst at 0.

\subsection{Future Scheme and Past Scheme}

\begin{figure}[!t]
\centering
\includegraphics[width=2.5in]{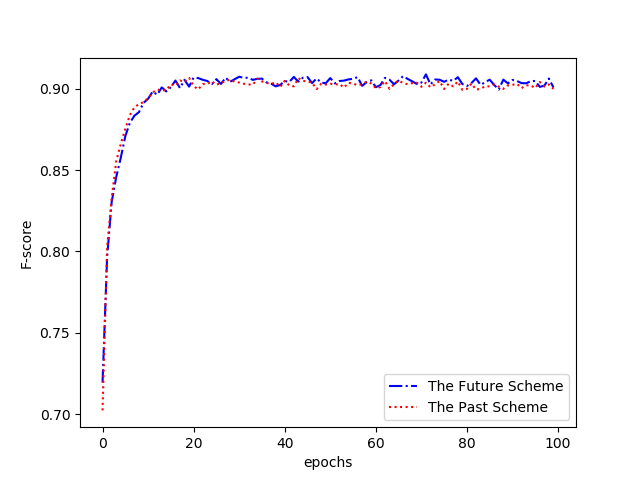}
\caption{The F-score for the models adopting the future scheme or the past scheme.}
\label{fig:fp-compare}
\end{figure}

\begin{table}[t!]
\begin{center}
\begin{tabular}{c|c|c|c}
\hline 
\textbf{Scheme} & \textbf{P} & \textbf{R} & \textbf{F} \\ 
\hline
The past scheme & 91.27 & 92.81 & 92.03 \\
\hline
\textbf{The future scheme} & \textbf{91.74} & \textbf{92.94} & \textbf{92.34} \\
\hline
\end{tabular}
\end{center}
\caption{Performance of the CWS comparison between the schemes}
\label{table:fp-comparison}
\end{table}

In generic temporal convolutional network, it focusses on the information at left scope in a sentence (the past scheme). However, in our model, we only focus on the information at the right scope in a sentence (the future scheme). We compare the difference between the schemes for convolutional scope in performance of the CWS on the test set of NLPCC2016. As shown in Table~\ref{table:fp-comparison}, this is a little difference between the schemes in performance, but the future scheme hits the better F score than the past scheme. The reason is that starting a dilated convolution from zero paddings by adopting the past scheme will lose the information of characters in front of a sentence. Instead, the future scheme is without the problem that is the loss of the information in front of a sentence. 

\subsection{Model Analysis}
\label{ssec:analysis}

\begin{figure}[!t]
\centering
\includegraphics[width=2.5in]{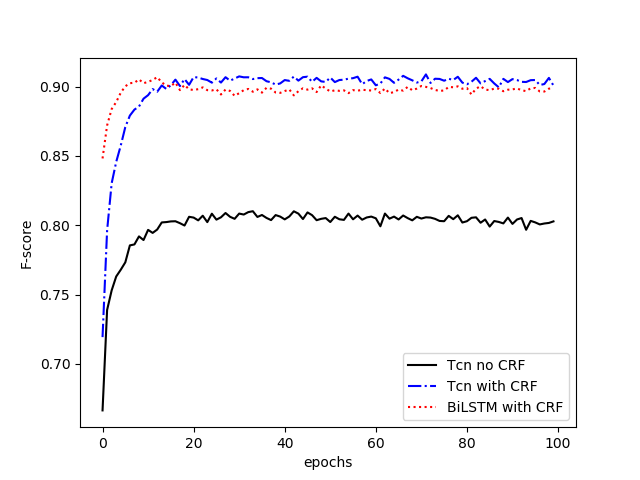}
\caption{The F-score for three models.}
\label{fig:fscore}
\end{figure}

The current state-of-the-art models for the task of Chinese word segmentation are mostly based the Bi-LSTM combining the CRF that can be named \textbf{Bi-LSTM-with-CRF}. The generic temporal convolutional network is without combining the CRF to train parameters and without using the Viterbi algorithm to infer true labels, which can be named \textbf{Tcn-no-CRF}. Our model has two important improvements contrasted the Tcn-no-CRF, combining the CRF to train parameters and using the Viterbi algorithm to infer true labels, and our model can be named \textbf{Tcn-with-CRF}. By the improvements, our model has a tremendous growth in performance of CWS contrasted the Tcn-no-CRF. By the dilated convolution, our model is without the correlation problem that is an inherent problem of the Bi-LSTM-with-CRF, so the power of parallel computing of our model is better than the Bi-LSTM-with-CRF. 

\begin{table}[t!]
\begin{center}
\begin{tabular}{c|c|c|c}
\hline 
\textbf{Model} & \textbf{F} & \textbf{Size} & \textbf{Speed} \\ 
\hline
Tcn-no-CRF & 82.89 & 2000 & 1287 \\
\hline
Bi-LSTM-with-CRF & 90.81 & 2080 & 120 \\
\hline
\textbf{Tcn-with-CRF} & \textbf{92.34} & \textbf{2000} & \textbf{263} \\
\hline
\end{tabular}
\end{center}
\caption{Performance comparison of the models. Size means the number of parameters of the models. Speed is the training speed of models and its unit of measurement is sentences/s meaning that how many sentences are processed per second.}
\label{table:comparison}
\end{table}

We select the Bi-LSTM-with-CRF and the Tcn-no-CRF as the baseline models, and experiment these models on NLPCC2016 dataset, and show the performance comparison between our model with the models on the test set. As shown in Figure~\ref{fig:fscore} and Table~\ref{table:comparison}, the Tcn-with-CRF boosts the performance of the CWS approximately 10\% contrasted the Tcn-no-CRF. The training speed is an important performance to evaluate the power of parallel computing in identical hardware and software environment. Meanwhile, if a product based on the algorithm of deep-learning is trained more quickly, the product is released more early before the other similar product. As shown in Table~\ref{table:comparison}, the training speed of the TCN-with-CRF has a tremendous growth than the Bi-LSTM-with-CRF, and the F-score of the TCN-with-CRF is better than the F-score of the Bi-LSTM-with-CRF. 

To better evaluate the performance of the CWS our model with other models premised the Bi-LSTM that other researchers proposed, we contrast the performance of the CWS between our model with prevalent models on the datasets of CITYU and PKU. As shown in Table~\ref{table:comparisons}, our model achieves the performance of the CWS equivalent to these models approximately. 

\begin{table*}[t!]
\centering
\begin{tabular}{|c|c|c|c|c|c|c|}
\hline
\multirow{2}{*}{Models} &
\multicolumn{3}{c|}{CITYU} &
\multicolumn{3}{c|}{PKU} \\
\cline{2-7}
 & P & R & F & P & R & F \\
\hline 
\cite{Zheng:2013} & - & - & - & 93.5 & 92.2 & 92.8 \\
\cite{Pei:2014} & - & - & - & 94.4 & 93.6 & 94.0 \\
\cite{Chen:2015} & - & - & - & 95.1 & 94.4 & 94.8 \\
\cite{Cai:2016} & - & - & - & 95.8 & 95.2 & 95.5 \\
\cite{Yang:2017} & - & - & 96.9 & - & - & 96.3 \\
\cite{Chen:2017} & 95.4 & 95.7 & 95.6 & 94.9 & 93.8 & 94.3 \\
\cite{Ma:2018} & - & - & 97.2 & - & - & 96.1 \\
\hline
\textbf{This work} & \textbf{93.8} & \textbf{94.4} & \textbf{94.1} & \textbf{94.6} & \textbf{94.2} & \textbf{94.4} \\
\hline
\end{tabular}
\caption{Comparison between our model with other prevalent models of the CWS.}
\label{table:comparisons}
\end{table*}

\section{Conclusions}
In this paper, we propose a sequence-to-sequence transformer model for the task of Chinese word segmentation. The model can apply the Dropout method to inhibit the Overfitting and can capture the character information at the distant site of a sentence by adding the layers of the encoder. Experiments show that our model has a tremendous growth in parallel computing than the models based on the Bi-LSTM, and our model achieves the performance of the CWS equivalent to the models based the Bi-LSTM approximately. However, our model gains a better of F-score than the baseline model based Bi-LSTM.     

\section*{Acknowledgments}

We would like to thank the anonymous reviewers for their valuable comments.  \\

\bibliography{acl2019}
\bibliographystyle{acl_natbib}

\end{document}